\def\paperlanguage{}
\pdfoutput=1 

\documentclass{tADR2e}

\usepackage{bm}
\usepackage{cite}
\include{preamble}

\newcommand{\ctext}[1]{\raise0.2ex\hbox{\textcircled{\scriptsize{#1}}}}
\definecolor{orange}{rgb}{1,0.5,0}
\definecolor{green}{rgb}{0,0.5,0}
\definecolor{lime}{rgb}{0,1.0,0}

\title{\LARGE \textbf
  {
    \switchlanguage%
    {%
      Analysis of Various Manipulator Configurations\\Based on Multi-Objective Black-Box Optimization
    }%
    {%
      多目的ブラックボックス最適化に基づく\\マニピュレータの最適コンフィギュレーションの分析
    }%
  }
}

\author{Kento Kawaharazuka$^{a}$$^{\ast}$, Keita Yoneda$^{a}$, Takahiro Hattori$^{a}$, Shintaro Inoue$^{a}$, and Kei Okada$^{a}$
    \thanks{$^\ast$Corresponding author. Email: kawaharazuka@jsk.imi.i.u-tokyo.ac.jp \vspace{6pt}}\\\vspace{6pt}  $^{a}${\em{The Department of Mechano-Informatics, Graduate School of Information Science and Technology, The University of Tokyo, 7-3-1 Hongo, Bunkyo-ku, Tokyo, Japan.}}
}
\begin{document}

\jvol{00} \jnum{00} \jyear{2025} \jmonth{December}

\maketitle

\begin{center}
  \small
  \switchlanguage%
  {%
    This is a preprint of an article whose final and definitive form has been published in ADVANCED ROBOTICS 2025, copyright Taylor \& Francis and Robotics Society of Japan, is available online at: http://www.tandfonline.com/Article DOI; https://doi.org/10.1080/01691864.2025.2607670.
  }%
  {%
    本稿はAdvanced Roboticsに投稿した査読前の原稿である.
  }%
\end{center}

\begin{abstract}
  \switchlanguage%
  {%
    Various 6-degree-of-freedom (DOF) and 7-DOF manipulators have been developed to date.
    Over a long history, their joint configurations and link length ratios have been determined empirically.
    In recent years, the development of robotic foundation models has become increasingly active, leading to the continuous proposal of various manipulators to support these models.
    However, none of these manipulators share exactly the same structure, as the order of joints and the ratio of link lengths differ among robots.
    Therefore, in order to discuss the optimal structure of a manipulator, we performed multi-objective optimization from the perspectives of end-effector reachability and joint torque.
    We analyze where existing manipulator structures stand within the sampling results of the optimization and provide insights for future manipulator design.
  }%
  {%
    今日まで様々な6自由度または7自由度のマニピュレータが開発されてきた.
    そして, 長い歴史の中で, それらの関節配置やリンク長さの比率は経験的に決定されてきた.
    特に現在はロボット基盤モデルの開発が盛んとなり, それを担う様々なマニピュレータが提案され続けている.
    しかし, そのどれもが全く同じ構造を持つわけではなく, 各ロボットで関節の順番やリンクの長さ比率は異なっている.
    そこで本研究では, 手先のreachabilityと関節トルクの観点から多目的最適化を行い, 最適なマニピュレータの構造について議論する.
    これまでのマニピュレータの構造が, その最適化サンプリングの中でどのような位置にあるのかを明らかにし, 今後のマニピュレータの設計に対して示唆を与える.
  }%
\end{abstract}


\section{Introduction}\label{sec:introduction}
\switchlanguage%
{%
  A wide range of 6-degree-of-freedom (DOF) and 7-DOF manipulators have been developed over the years.
  These manipulators have a long history, beginning with the development of Unimate \cite{devol1961unimate}, the world's first industrial manipulator, in 1961.
  Since then, many manipulators have been proposed, leading up to today's ALOHA \cite{zhao2023aloha} for robotic foundation models \cite{bommasani2021foundation, firoozi2024foundation, kawaharazuka2024foundation}.
  Unimate has joints arranged in the order of YPSPY (where Roll is denoted as R, Pitch as P, Yaw as Y, and Slide as S, and the sequence is described from the base of the manipulator. The initial state of a serial manipulator is defined as the configuration in which all links are extended vertically upward, and the joint types (R, P, Y, S) are determined based on the direction of movement in this state.)
  Although some robots have been developed using slide joints, most subsequent general-purpose manipulators have been composed of Roll, Pitch, and Yaw joints.
  Among 6-DOF manipulators, ALOHA \cite{zhao2023aloha}, COBOTTA \cite{denso2025cobotta}, and xArm 6 \cite{ufactory2025xarm6} have joints arranged in the order of YPPYPY.
  Additionally, the ARX \cite{arx2025robot} used in the recent $\pi_0$ \cite{black2024pi0} has a joint order of YPPPRY, while myCobot \cite{elephant2025mycobot} follows the order of YPPPYR.
  For 7-DOF manipulators, Franka Emika Panda \cite{frankaemika2025panda} has a joint order of YPYPYPR, while Sawyer \cite{rethink2025sawyer} and Kinova Gen3 \cite{kinova2025gen3} have joints arranged in the order of YPYPYPY.
  As seen above, the order of joints in manipulators varies from robot to robot, and the ratio of link lengths also differs accordingly.
}%
{%
  これまで様々な6自由度または7自由度のマニピュレータが開発されてきた.
  それらには古い歴史があり, 世界初の産業用マニピュレータであるUnimate \cite{devol1961unimate}が1961年に開発されて以来, 今日のALOHA \cite{zhao2023aloha}に至るまで, 多くのマニピュレータが提案されている.
  Unimate \cite{devol1961unimate}はYPSPYの順番で関節が配置されている(なお, RollをR, PitchをP, YawをY, SlideをSとして表記し, マニピュレータの根本から順に記述する. また, シリアルマニピュレータのリンクが全てが鉛直上向きに伸びた状態を初期状態とし, その状態での関節の動く向きからR,P, Y, Sを決めている.)
  直動関節を使用したロボットはいくつか提案されているが, その後の一般的なマニピュレータはRoll, Pitch, Yaw関節により構成されている.
  6自由度だと, ALOHA \cite{zhao2023aloha}やCOBOTTA \cite{denso2025cobotta}, xArm 6 \cite{ufactory2025xarm6}などのマニピュレータはYPPYPYの順番で関節が配置されている.
  また, 近年の$\pi_0$ \cite{black2024pi0}に用いられているARX \cite{arx2025robot}はYPPPRYの順番, myCobot \cite{elephant2025mycobot}はYPPPYRの順番である.
  他にも, 7自由度だと, Franka Emika Panda \cite{frankaemika2025panda}はYPYPYPR, Sawyer \cite{rethink2025sawyer}やKinova Gen3 \cite{kinova2025gen3}はYPYPYPYの順番で関節が配置されている.
  このように, マニピュレータの関節の順番はロボットごとに異なっており, また, リンク長さの比率も異なる.
}%

\begin{figure}[t]
  \centering
  \includegraphics[width=0.7\columnwidth]{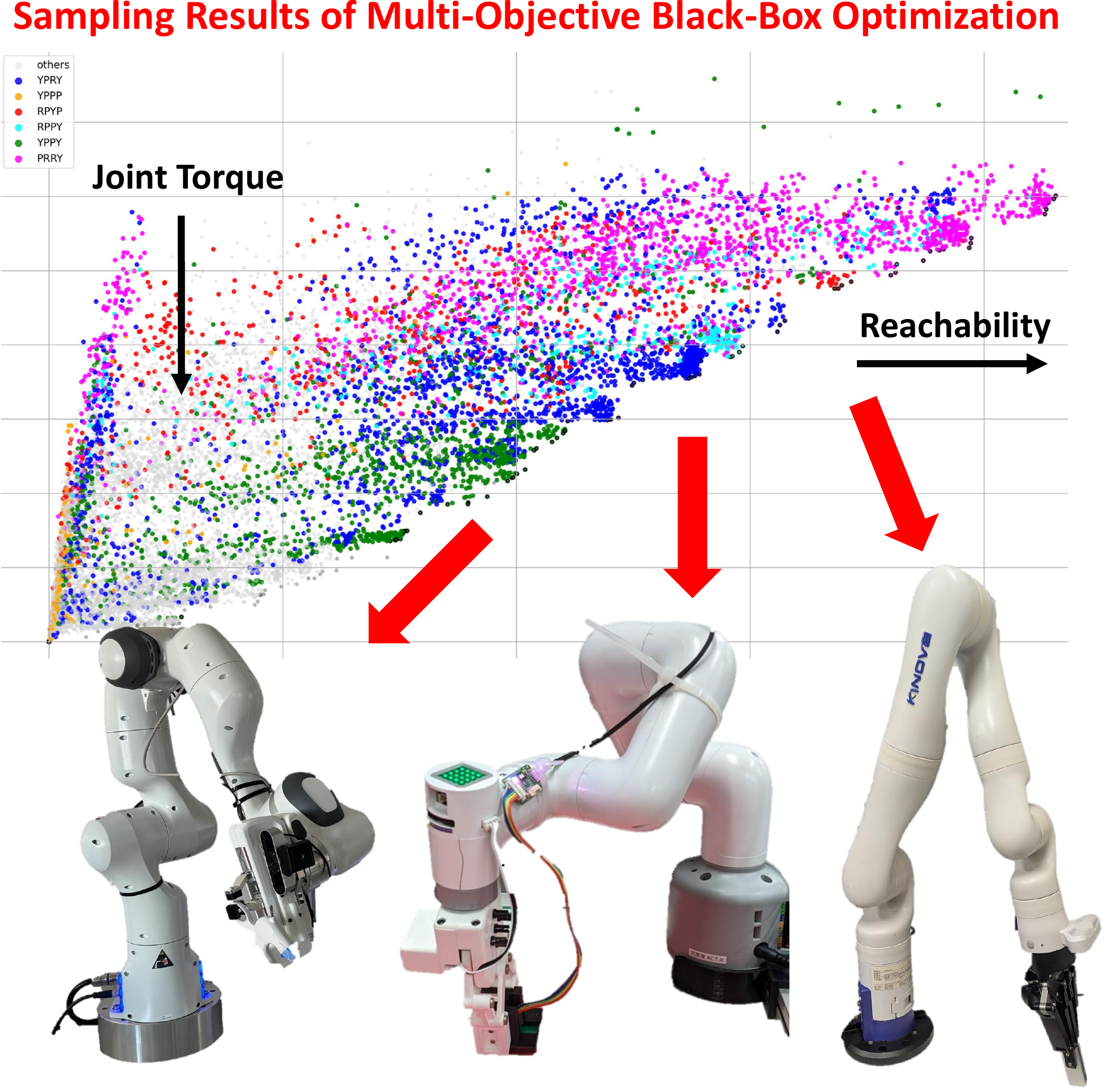}
  \caption{Research concept: By performing multi-objective optimization from the perspectives of end-effector reachability and joint torque, this study discusses various joint structures, including the diverse manipulators that have been actively developed in recent years.}
  \label{figure:concept}
\end{figure}

\switchlanguage%
{%
  The structure of manipulators has traditionally been determined empirically.
  However, as various manipulator configurations continue to be developed, numerous studies have focused on optimizing body structure design to determine the most effective configurations.
  In \cite{paden1988optimal}, an optimal joint configuration for a 6-DOF manipulator was proposed based on workspace maximization and the concept of a well-connected workspace.
  In \cite{hollerbach1985optimum}, heuristic methods were used to extract several joint configurations for a 7-DOF manipulator, which were then evaluated based on singularity avoidance, workspace optimization, kinematic simplicity, and mechanical feasibility.
  For industrial robots, \cite{yang2000modular} optimized the number and types of modules, as well as their relative positions, using a genetic algorithm to satisfy desired task points.
  In \cite{xiao2016designopt}, a multi-objective optimization of motor and gear ratios for a 6-DOF manipulator was performed using a genetic algorithm, aiming to minimize body weight while maximizing manipulability.
  In \cite{liu2020modular}, an exhaustive search was conducted to optimize the design of modular robots capable of executing tasks while satisfying constraints such as joint angle limits, torque limits, and collision avoidance.
  In \cite{lei2024optimization}, an evolutionary strategy was used to optimize module placement to minimize required torque and maximize manipulability while following a commanded trajectory.
  Furthermore, \cite{kawaharazuka2023autodesign, kawaharazuka2024slideopt} proposed a Pareto-optimal solution for minimizing control error and required torque through multi-objective optimization using a genetic algorithm, enabling body structure adaptation based on tasks and user preferences.

  These studies can be categorized based on different approaches.
  For example, there are studies utilizing analytical optimization \cite{hollerbach1985optimum, paden1988optimal} and black-box optimization \cite{yang2000modular, xiao2016designopt, liu2020modular, lei2024optimization, kawaharazuka2023autodesign}.
  Additionally, there are studies focused on evaluating task-independent universal performance \cite{paden1988optimal, hollerbach1985optimum} and those that perform task-specific optimization \cite{yang2000modular, xiao2016designopt, liu2020modular, lei2024optimization, kawaharazuka2023autodesign}.
  Historically, analytical methods were used to optimize task-independent performance \cite{paden1988optimal, hollerbach1985optimum}.
  On the other hand, recent research has increasingly focused on performing more complex and multi-objective optimizations tailored to specific tasks using black-box optimization methods \cite{yang2000modular, xiao2016designopt, liu2020modular, lei2024optimization, kawaharazuka2023autodesign}.
  Analytical optimization have limitations, such as the inability to account for joint angle limits and link collisions.
  Also, they only focus on reachability, without considering multi-objective optimization which includes joint torque.
  In contrast, modern black-box optimization methods enable the consideration of multiple constraints and multi-objective optimization.
  However, most of these studies focus on task-specific optimization, and there has been little effort to perform a unified multi-objective optimization of general manipulator structures, providing a comprehensive discussion of the diverse manipulator structures currently being developed.

  To address this gap, this study performs a multi-objective optimization of general 6-DOF and 7-DOF manipulators from the perspectives of end-effector reachability and joint torque, analyzing various manipulator structures comprehensively (\figref{figure:concept}).
  The optimization considers discrete joint types and continuous link lengths as parameters, applying certain constraints and using a Tree-Structured Parzen Estimator (TPE) \cite{bergstra2011tpe} for multi-objective optimization.
  This approach clarifies the position of existing manipulator structures within the optimization sampling and provides insights for future manipulator design.

  The remainder of this paper is structured as follows.
  \secref{sec:proposed} describes the design parameters, constraints, objective functions, and details of the multi-objective optimization used in this study.
  \secref{sec:experiment} presents the results of the multi-objective optimization for 6-DOF and 7-DOF manipulators and discusses the findings based on the sampling results for currently developed manipulators.
  \secref{sec:discussion} provides further discussion of the study's results, and conclusions are presented in \secref{sec:conclusion}.
}%
{%
  これまで, マニピュレータの構造は経験的に決定されてきた.
  一方で, 様々な構成のマニピュレータが構築されるにしたがって, どのような構成が良いのか, 身体構造の設計を最適化する研究も多数存在している.
  \cite{paden1988optimal}では, 6自由度マニピュレータについて, workspaceの最大化とwell-connectd workspaceの考え方に基づく最適な関節配置を提案している.
  \cite{hollerbach1985optimum}では, 7自由度マニピュレータについて, ヒューリスティックにいくつかの関節配置を抽出し, それらを特異点の排除・作業空間の最適化・運動学的な単純性・機械的な実現可能性の観点から評価している.
  \cite{yang2000modular}では, 産業用ロボットに向け, 所望の動作点を満たすモジュールの数と種類, モジュール間の相対的な位置を遺伝的アルゴリズムにより最適化した.
  \cite{xiao2016designopt}では, 身体重量の最小化とmanipulabilityの最大化に基づき, 6自由度マニピュレータのモータとギア比を遺伝的アルゴリズムにより多目的最適化した.
  \cite{liu2020modular}では, 総当りにより関節角度制限やトルク制限, 衝突回避等を満たしつつタスクを実行可能なモジュールロボットの設計最適化を行った.
  \cite{lei2024optimization}では, 指令された軌道を満たしつつmanipulabilityと必要トルクを最小化するモジュール配置を進化戦略により最適化した.
  \cite{kawaharazuka2023autodesign}では, タスクやユーザの好みに応じて身体を変化させるために, 遺伝的アルゴリズムを用いた多目的最適化により制御誤差と必要トルクを最小化するパレート解を提示している.

  これらはいくつかの観点から分類できる.
  例えば, 解析的最適化\cite{hollerbach1985optimum, paden1988optimal}とブラックボックス最適化\cite{yang2000modular, xiao2016designopt, liu2020modular, lei2024optimization, kawaharazuka2023autodesign}が存在している.
  また, タスクに依存しない普遍的な性能の評価\cite{paden1988optimal, hollerbach1985optimum}や, タスクに応じた最適化\cite{yang2000modular, xiao2016designopt, liu2020modular, lei2024optimization, kawaharazuka2023autodesign}が存在している.
  古くはタスクに依存しない普遍的な性能の最適化を解析的に最適化することが行われてきたが\cite{paden1988optimal, hollerbach1985optimum}, 現在はタスクに応じたより複雑で多目的な最適化を, ブラックボックス最適化により行う研究が盛んである\cite{yang2000modular, xiao2016designopt, liu2020modular, lei2024optimization, kawaharazuka2023autodesign}.
  解析的な最適化は, 関節角度リミットやリンク間衝突が考慮できない, 手先の到達範囲しか考えられないため関節トルク等と合わせた多目的最適化ができないなど, 多くの制約がある.
  これに対して, 近年のブラックボックス最適化を活用した議論では, 様々な制約の考慮や多目的最適化が可能である.
  その一方で, タスクに特化した最適化が多く, 汎用的なマニピュレータの構造を多目的最適化し, 現在盛んに開発されている多様なマニピュレータの構造を統一的に議論することが行われてきていない.

  そこで本研究では, 6自由度または7自由度のマニピュレータについて, 手先のreachabilityと関節トルクの観点から多目的最適化を行い, 多様なマニピュレータの構造について議論する(\figref{figure:concept}).
  離散的な関節の種類と連続的なリンクの長さをパラメータとし, いくつかの制約を加えた上で, Tree-Structured Parzen Estimator (TPE)を用いた多目的最適化を行う.
  これにより, これまでのマニピュレータの構造が, その最適化サンプリングの中でどのような位置にあるのかを明らかにし, 今後のマニピュレータの設計に対する示唆を与える.

  本研究の構成は以下の通りである.
  \secref{sec:proposed}では, 本研究で考える設計パラメータと制約条件, 最適化する評価関数, 多目的最適化の詳細について述べる.
  \secref{sec:experiment}では, 6自由度または7自由度のマニピュレータについて多目的最適化を行い, 現在開発されている多様なマニピュレータについて, サンプリング結果から議論を行う.
  \secref{sec:discussion}では, 本研究の結果について議論し, \secref{sec:conclusion}で結論を述べる.
}%

\begin{figure}[t]
  \centering
  \includegraphics[width=1.0\columnwidth]{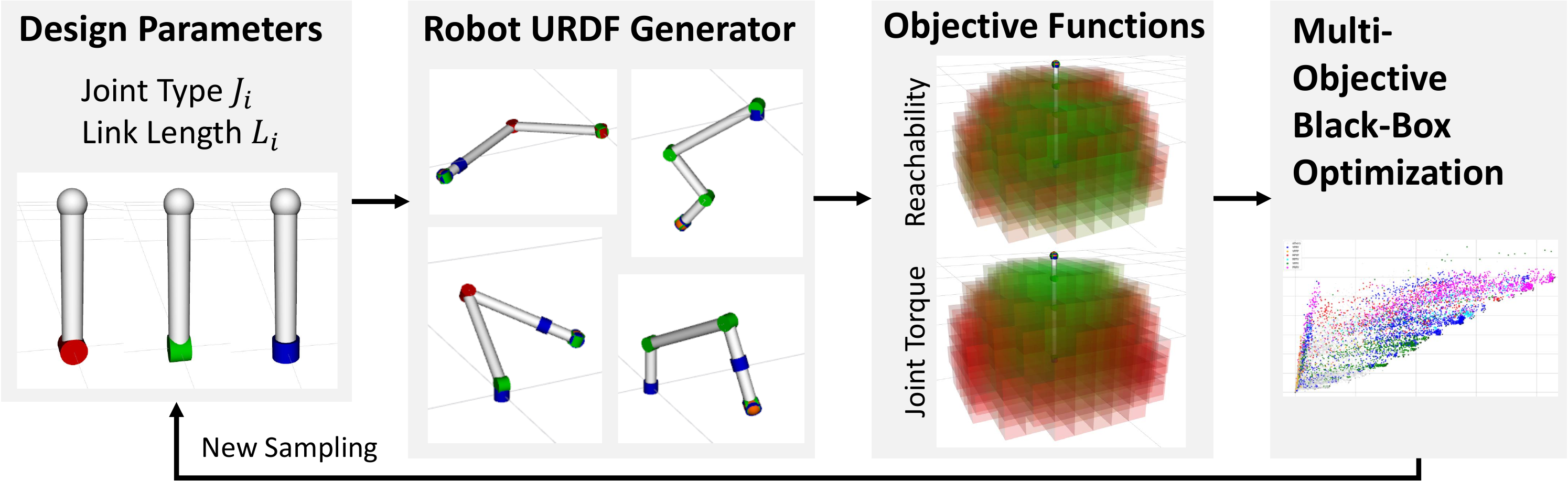}
  \caption{System overview: This study parametrizes joint types and link lengths, automatically generates URDFs, and performs multi-objective optimization from the perspectives of end-effector reachability and joint torque to discuss the structures of various manipulators.}
  \label{figure:overview}
\end{figure}

\section{Design of Various Manipulator Configurations Based on Multi-Objective Black-Box Optimization} \label{sec:proposed}
\switchlanguage%
{%
  The overall framework of this study is shown in \figref{figure:overview}.
  First, the design parameters for joint types and link lengths are determined.
  Second, individual joints described as xacro (XML macro) files are combined to automatically generate the robot's URDF (Unified Robot Description Format).
  Finally, the objective functions -- end-effector reachability and joint torque -- are computed, and multi-objective optimization is performed based on these criteria.
}%
{%
  本研究の全体像を\figref{figure:overview}に示す.
  まず関節の種類とリンク長の設計パラメータを決定する.
  次に, xacro (XML macro)ファイルとして記述された各関節を組み合わせ, ロボットのURDF (Unified Robot Description Format)を自動的に構築する.
  最後に, 目的関数として手先のreachabilityと関節トルクを計算し, これらをもとに多目的最適化を実行する.
}%

\subsection{Design Parameters of Manipulators} \label{subsec:parameters}
\begin{figure}[t]
  \centering
  \includegraphics[width=0.7\columnwidth]{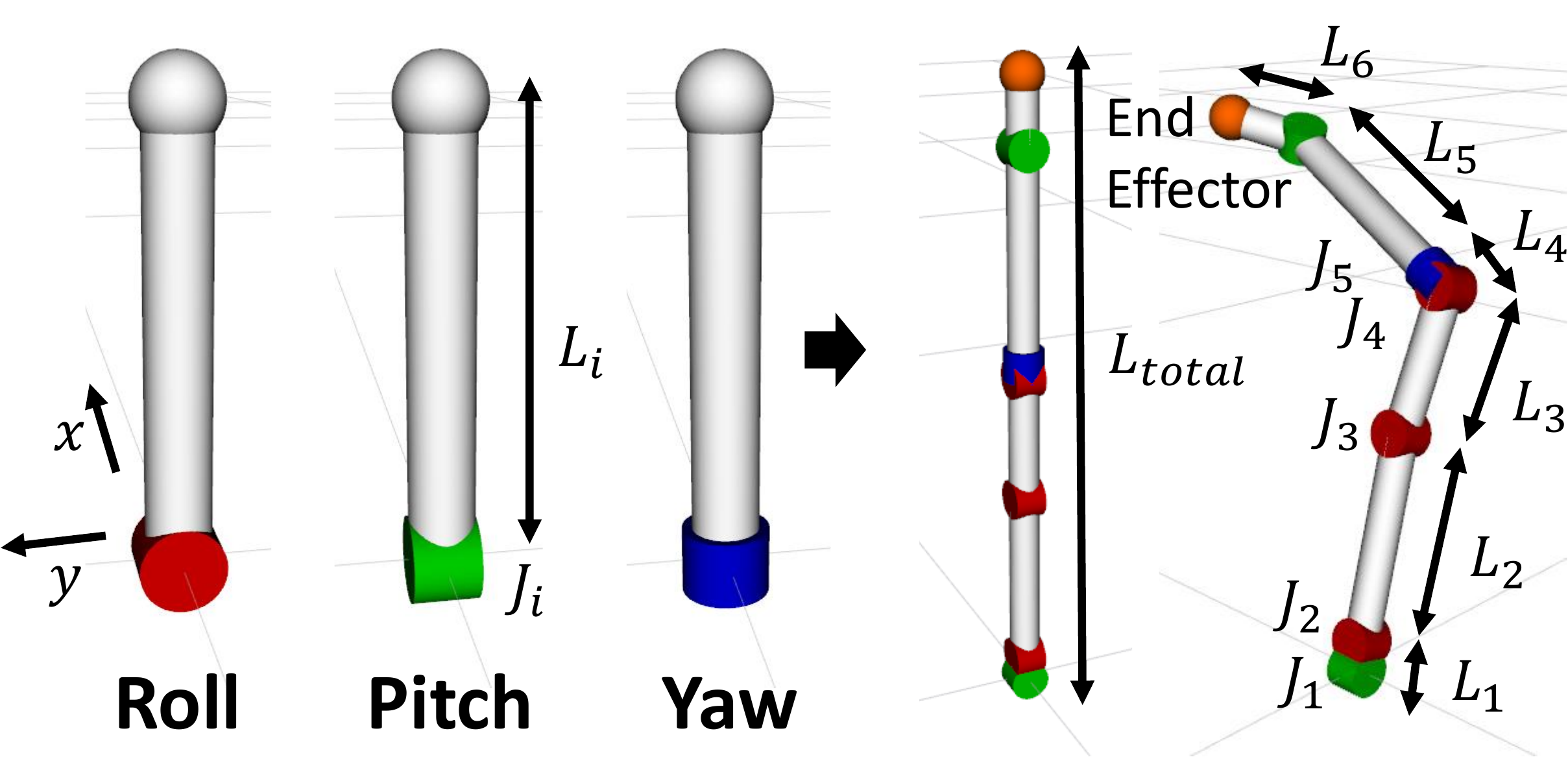}
  \caption{Design parameters handled in this study: The joint type $J_i$ and link length $L_i$ are varied as design parameters.}
  \label{figure:params}
\end{figure}

\switchlanguage%
{%
  The design parameters considered in this study are described in \figref{figure:params}.
  Here, the origin of the manipulator coincides with the origin of the workspace.
  Additionally, in the initial posture where all joint angles are set to $0$, the manipulator extends straight upward in the vertical direction.
  The types of joint angles are determined based on this initial posture.

  For an $N_{joint}$-DOF manipulator, the joint types and link lengths are treated as design parameters.
  Each joint $J_i$ ($1 \leq i \leq N_{joint}$) is selected from three types: Roll, Pitch, and Yaw.
  The link length $L_i$ ($1 \leq i \leq N_{joint}$) is represented as a value within the range $[0, L^{max}]$.
  The URDF of the manipulator is constructed by including the xacro files of each joint.
  Each joint's xacro file contains parameters such as the link length following that joint, motor mass, and link mass.
  The motor mass is denoted as $m_{motor}$, while the link mass is expressed as $\pi r^2 L_i \rho$, where $r$ represents the link radius and $\rho$ denotes the link density.
  Furthermore, the joint angle limits for all joints are set within the range $[-\frac{3}{4}\pi, \frac{3}{4}\pi]$.

  Several constraints are imposed on these design parameters.
  First, consecutive Yaw joints are not allowed, as having a Yaw joint followed by another Yaw joint is redundant.
  Next, a Roll or Pitch joint following a Yaw joint can achieve the same motion by rotating the Yaw joint.
  Therefore, these cases are unified into Pitch joints.
  In summary, if $J_i$ is a Yaw joint, then $J_{i+1}$ must be a Pitch joint.
  Additionally, the total link length, i.e., the overall length of the manipulator, is fixed at $L_{total}$.
  The specific implementation of these constraints is discussed in \secref{subsec:mobbo}.
}%
{%
  本研究で扱う設計パラメータについて述べる(\figref{figure:params}).
  前提として, このマニピュレータの原点と作業空間の原点は一致している.
  また, 関節角度が全て$0$の初期姿勢において, このマニピュレータは鉛直上向き方向に真っ直ぐ伸びているものとし, 関節角度の種類はこの初期姿勢を基準に決定される.

  $N_{joint}$自由度のマニピュレータについて, 各関節の種類とリンクの長さを設計パラメータとする.
  各関節$J_i$ ($1 \leq i \leq N_{joint}$)はRoll, Pitch, Yawの3種類の中から選択され, リンクの長さ$L_i$ ($1 \leq i \leq N_{joint}$)は$[0, L^{max}]$の間の数値として表現される.
  各関節のxacroファイルをincludeすることでマニピュレータのURDFを構築する.
  各関節のxacroファイルには, その関節の後に続くリンク長さ, モータ重量, リンク重量のパラメータが含まれる.
  モータの重量を$m_{motor}$, リンクの重量を$\pi r^2 L_i \rho$とする($r$はリンクの半径, $\rho$はリンクの密度を表す).
  また, 関節角度限界は全ての関節で$[-\frac{3}{2}\pi, \frac{3}{2}\pi]$の範囲とする.

  この設計パラメータに対してはいくつかの制約を設ける.
  まず, Yaw関節の後にYaw関節が続いても意味がないため, このような関節配置は許容しない.
  次に, Yaw関節の後に続くRoll関節とPitch関節は, Yaw関節を回転させることで同じ動作が可能であるため, これらはPitch関節に統一する.
  つまりまとめると, $J_{i}$がYawのとき, $J_{i+1}$は必ずPitch関節となる.
  また, リンク長さの合計, つまりマニピュレータの全長は$L_{total}$に固定する.
  これらの制約の具体的な考慮の方法は\secref{subsec:mobbo}で述べる.
}%

\subsection{Objective Functions of Reachability and Joint Torque} \label{subsec:objective}
\begin{figure}[t]
  \centering
  \includegraphics[width=0.9\columnwidth]{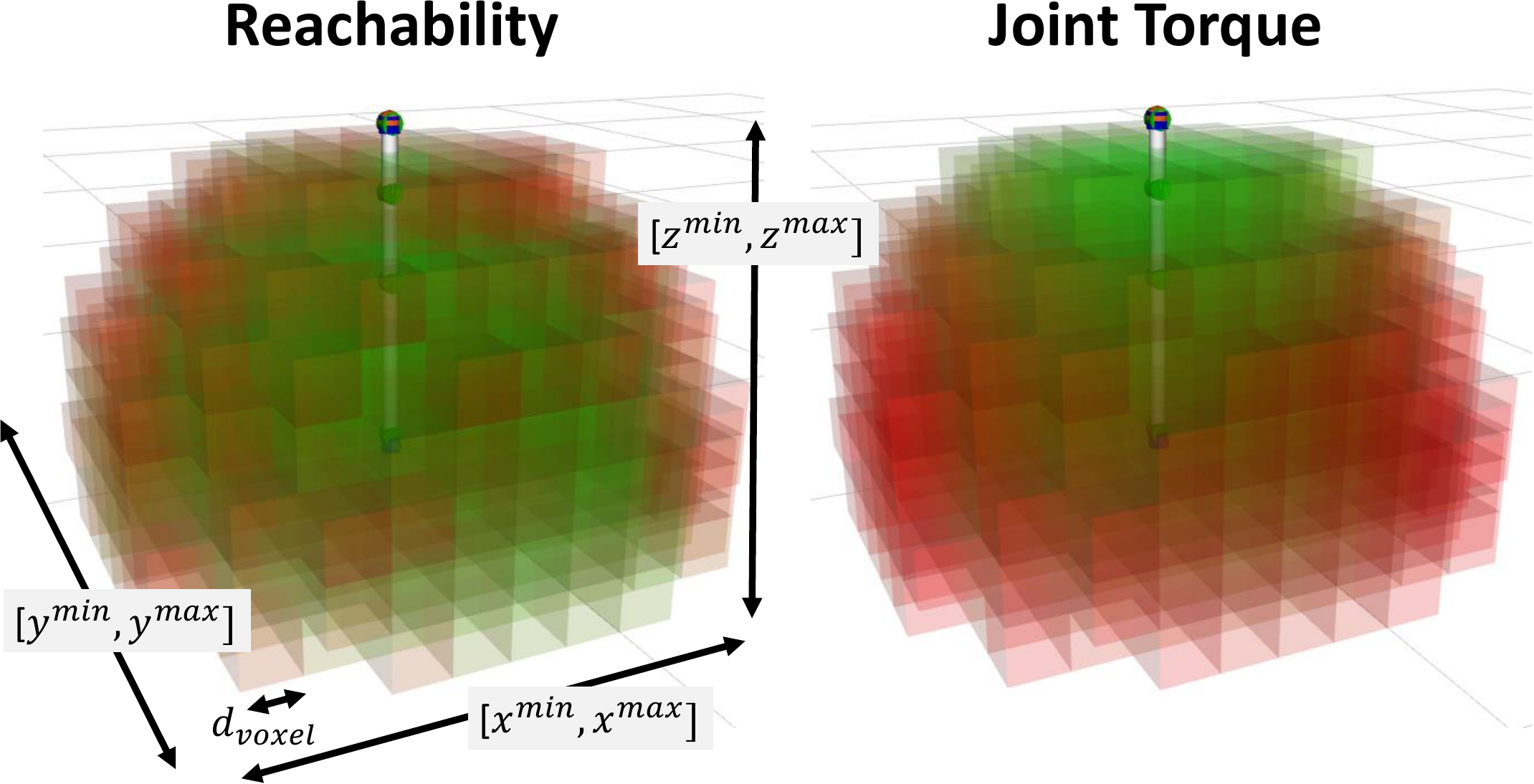}
  \caption{Objective functions for multi-objective optimization: The left image shows end-effector reachability, while the right image shows necessary joint torque.}
  \label{figure:objective}
\end{figure}

\switchlanguage%
{%
  The objective functions used in the optimization process of this study are described below.
  Here, a multi-objective optimization is performed based on two criteria: end-effector reachability \cite{zacharias2007reachability} and joint torque (\figref{figure:objective}).
  Of course, it is possible to consider additional perspectives, but due to the difficulty of visualizing them and the fact that the evaluation function defined in this study -- based on position and force -- is the most general-purpose, the optimization is conducted with respect to these two aspects.

  First, the workspace for evaluation is defined as follows.
  \begin{equation}
    x \in [x^{\min}, x^{\max}], \quad y \in [y^{\min}, y^{\max}], \quad z \in [z^{\min}, z^{\max}]
  \end{equation}
  This workspace is uniformly voxelized with an interval of $d_{voxel}$, and the set of all voxels is denoted as $\mathcal{V}$.
  For each voxel $i \in \mathcal{V}$, let $\bm{p}_{i}$ be the center position of that voxel.
  Additionally, for each voxel $i$, $N_{rand}$ random end-effector postures are generated, and inverse kinematics is solved while changing the posture at position $\bm{p}_{i}$.
  The success rate of the inverse kinematics solution is defined as the reachability score $E^{reach}_{i}$ and is formulated as follows,
  \begin{equation}
    E^{reach}_{i} = \frac{\sum_{j=1}^{N_{rand}} S_{ij}}{N_{rand}}
  \end{equation}
  where $S_{ij} = 1$ if the inverse kinematics problem is successfully solved for the $j$-th random posture in voxel $i$, and $S_{ij} = 0$ otherwise.
  The inverse kinematics algorithm used follows the methods proposed in \cite{chan1995ik, sugiura2007ik}, which also account for joint angle limits and collision detection between links.
  Due to the dependence on initial conditions, if the inverse kinematics solution fails, it is retried up to $N_{ik}=5$ times with random initial joint angles.
  Additionally, for each voxel $i$, the norm of the average required joint torque $E^{torque}_{i}$ for successful inverse kinematics solutions is computed as follows,
  \begin{equation}
    E^{torque}_{i} = \frac{1}{\sum_{j=1}^{N_{rand}} S_{ij}} \sum_{j=1}^{N_{rand}} S_{ij} \| \bm{\tau}_{ij} \|
  \end{equation}
  where $\bm{\tau}_{ij} \in \mathbb{R}^{N_{joint}}$ is the joint torque vector required for the $j$-th solution in voxel $i$, and $|| \cdot ||$ represents the L2 norm.
  Finally, the overall reachability and joint torque scores across the entire workspace are computed as follows.
  \begin{equation}
    E^{reach} = \sum_{i \in \mathcal{V}} E^{reach}_{i}, \quad E^{torque} = \sum_{i \in \mathcal{V}} E^{torque}_{i}
  \end{equation}

  For example, \figref{figure:objective} illustrates the reachability and joint torque distribution for a manipulator with a YPPPYP joint structure.
  The parameters were set as follows: $x^{\min} = -0.5, x^{\max} = 0.5, y^{\min} = -0.5, y^{\max} = 0.5, z^{\min} = -0.1, z^{\max} = 0.5, d_{voxel} = 0.1$, and $N_{rand} = 30$.
  For each voxel in the visualization, reachability is represented with red for $E^{reach}_{i}=0$ and green for $E^{reach}_{i}=1$.
  Also, joint torque is represented with green for $E^{torque}_{i}=3.0$ Nm and red for $E^{torque}_{i}=10.0$ Nm (this visualization approach will also be used in subsequent experiments).
  In this example, reachability is higher near the origin and decreases as the distance from the origin increases.
  Additionally, joint torque decreases as the manipulator extends in the $z$-direction and increases as it extends in the $x$- or $y$-direction.

}%
{%
  本研究の最適化で扱う評価関数について述べる.
  ここでは, 手先のreachability \cite{zacharias2007reachability}と関節トルクの2つの観点から多目的最適化を行う(\figref{figure:objective}).
  なお, もちろんさらに複数の観点を考慮することも可能であるが, 描画が難しいこと, 今回設定した評価関数が位置と力という点で最も汎用的であることから, この2つの観点で最適化を行っている.

  まず, 評価を行う作業空間の範囲を以下のように設定する.
  \begin{equation}
    x \in [x^{\min}, x^{\max}], \quad y \in [y^{\min}, y^{\max}], \quad z \in [z^{\min}, z^{\max}]
  \end{equation}
  この空間を一様に$d_{voxel}$の間隔でボクセル化し, 全体のボクセルの集合$\mathcal{V}$を定義する.
  各ボクセル$i \in \mathcal{V}$に対して, そのボクセルの中心位置を$\bm{p}_i$とする.
  また, 各ボクセル$i$に対して, $N_{rand}$個のランダムな手先姿勢を生成し, 位置$\bm{p}_i$対して姿勢を変更しながら逆運動学を解く.
  逆運動学の成功率を手先のreachabilityスコア$E^{reach}_{i}$とし, 以下のように定義する.
  \begin{equation}
    E^{reach}_{i} = \frac{\sum_{j=1}^{N_{rand}} S_{ij}}{N_{rand}}
  \end{equation}
  ここで,ボクセル$i$において$j$番目のランダムな姿勢に対して逆運動学問題を解くことに成功した場合$S_{ij} = 1$, 解けなかった場合$S_{ij} = 0$とする.
  なお, 逆運動のアルゴリズムは\cite{chan1995ik, sugiura2007ik}におけるアルゴリズムを用いており, 関節角度限界やリンク間の衝突判定も行っている.
  初期値依存性があるため, 逆運動学が解けなかった場合, $N_{ik}=5$回まで初期関節角度をランダムに変更しながら再試行する.
  加えて, 各ボクセル$i$において, 成功した逆運動学解に必要な関節トルクのノルム平均$E^{torque}_{i}$を計算する.
  \begin{equation}
    E^{torque}_{i} = \frac{1}{\sum_{j=1}^{N_{rand}} S_{ij}} \sum_{j=1}^{N_{rand}} S_{ij} \| \bm{\tau}_{ij} \|
  \end{equation}
  ここで, $\bm{\tau}_{ij} \in \mathbb{R}^{N_{joint}}$はボクセル$i$において$j$番目の解に必要な関節トルクベクトル, $\| \cdot \|$はL2ノルムを表す.

  最後に, 全体の作業空間に対するreachabilityと関節トルクのスコアを以下のように計算する.
  \begin{equation}
    E^{reach} = \sum_{i \in \mathcal{V}} E^{reach}_{i}, \quad E^{torque} = \sum_{i \in \mathcal{V}} E^{torque}_{i}
  \end{equation}

  例えば, \figref{figure:objective}に示したのは, YPPPYPの関節構造を持つマニピュレータの手先のreachabilityと関節トルクの例である.
  $x^{\min} = -0.5, x^{\max} = 0.5, y^{\min} = -0.5, y^{\max} = 0.5, z^{\min} = -0.1, z^{\max} = 0.5, d_{voxel} = 0.1, N_{rand} = 30$とした.
  各ボクセルについて, reachabilityでは$E^{reach}_{i}=0$のときに赤, $E^{reach}_{i}=1$のときに緑となり, 関節トルクでは$E^{torque}_{i}=3.0$ Nmで緑, $E^{torque}_{i}=10.0$ Nmで赤となるようにグラデーションをつけ描画している(今後の実験も同様である).
  この例では, 原点から近いところでreachabilityが高く, 原点から遠いほどreachabilityが低い.
  また, $z$方向に伸びるほど関節トルクが小さく, $z$方向に縮むほど関節トルクが大きいという特性がわかる.
}%

\subsection{Multi-Objective Black-Box Optimization} \label{subsec:mobbo}
\switchlanguage%
{%
  Based on the previously described design parameters and objective functions, multi-objective optimization is performed.
  For the multi-objective optimization, the Tree-Structured Parzen Estimator (TPE) \cite{bergstra2011tpe} is used.
  Specifically, the multivariate TPE \cite{falkner2018multivariate}, which considers dependencies between parameters, is employed.
  This enables the discovery of various design solutions from the sampling results.

  Here, $J_i$ is treated as a categorical variable with three choices, while $L_i$ is treated as a continuous variable.
  The optimization is performed based on maximizing $E^{reach}$ and minimizing $E^{torque}$.
  During this process, the constraints described in \secref{subsec:parameters} must be considered.
  First, to enforce the rule that a Pitch joint follows a Yaw joint, when $J_i$ is Yaw, $J_{i+1}$ is set to Pitch regardless of which choice is selected.
  This is due to the limitation in the optuna \cite{akiba2019optuna} implementation, where categorical variable choices cannot be dynamically modified.
  Next, since the total link length is fixed at $L_{total}$, $L^{max}$ is dynamically adjusted regarding $L_i$.
  For $i=1$, $L^{max}$ is set as $L^{max} = L^{max}_{init}$ (where $L^{max}_{init}$ is a constant).
  For $i>1$, $L^{max}$ is updated dynamically as follows.
  \begin{align}
    L^{remain}_{i} = L_{total} - \sum_{j=1}^{i-1} L_j \\
    L^{max}_{i} = \min(L^{max}_{init}, L^{remain}_{i})
  \end{align}
  Furthermore, to ensure the total link length remains $L_{total}$, the final link length is set as $L_{N_{joint}}=L^{remain}_{N_{joint}}$.
  Thus, the actual parameters optimized are $J_i$ ($1 \leq i \leq N_{joint}$) and $L_i$ ($1 \leq i \leq N_{joint}-1$).
  For other parameters, the number of initial random trials for multi-objective optimization is set to $N^{trial}_{init}=100$, and the total number of optimization trials is set to $N^{trial}_{total}=20000$.
}%
{%
  これまでに述べた設計パラメータと評価関数に基づき, 多目的最適化を行う.
  多目的最適化にはTree-Structured Parzen Estimator (TPE) \cite{bergstra2011tpe}を用いる.
  その中でも特に, multivariate TPE \cite{falkner2018multivariate}と呼ばれる, パラメータ間の依存関係を考慮したTPEを用いている.
  これによって, サンプリング結果から様々な設計解を得ることができる.

  ここでは, $J_i$を3つの選択肢を持つカテゴリカル変数, $L_i$を連続変数として扱い, $E^{reach}$の最大化, $E^{torque}$の最小化に基づき最適化を行う.
  この際, \secref{subsec:parameters}で述べた制約を考慮する必要がある.
  まず, Yaw関節の後をPitch関節に統一するため, $J_i$がYawのとき, $J_{i+1}$自体は3つの選択肢から選ばれるが, 実際に用いられる関節はPitch関節とする.
  これは, optuna \cite{akiba2019optuna}の実装の都合上, カテゴリカル変数の選択肢を動的に変更できないためである.
  次に$L_i$については, リンク長さの合計が$L_{total}$に固定されるため, $L^{max}$を動的に変更する.
  $i=1$のときは$L^{max} = L^{max}_{init}$ ($L^{max}_{init}$は定数)とし, $i>1$のときは以下のように$L^{max}=L^{max}_{i}$とする.
  \begin{align}
    L^{remain}_{i} = L_{total} - \sum_{j=1}^{i-1} L_j \\
    L^{max}_{i} = \min(L^{max}_{init}, L^{remain}_{i})
  \end{align}
  また, 全体のリンク長さを$L_{total}$とするため, $L_{N_{joint}}=L^{remain}_{N_{joint}}$とする.
  つまり, 実際に最適化されるパラメータは$J_i$ ($1 \leq i \leq N_{joint}$)と$L_i$ ($1 \leq i \leq N_{joint}-1$)となる.

  その他のパラメータについては, 多目的最適化の初期ランダム試行を$N^{trial}_{init}=100$, 全体の最適化試行を$N^{trial}_{total}=20000$とした.
}%

\section{Experiments} \label{sec:experiment}

\switchlanguage%
{%
  The experimental setup of this study is described below.
  First, two experiments are conducted by varying the number of joints as $N_{joint}=\{6,7\}$.
  For the parameters related to the objective function calculation, the following values are used considering generality and computational cost: $x^{\min} = -0.5, x^{\max} = 0.5, y^{\min} = -0.5, y^{\max} = 0.5, z^{\min} = -0.1, z^{\max} = 0.5, d_{voxel} = 0.2$, and $N_{rand} = 30$.
  For other parameters, the following values are set considering general manipulator configurations: $m_{motor}=0.5$ kg, $\rho=1.0\times10^3$ kg/m$^3$, $r=0.015$ m, $L_{total}=0.6$ m, and $L^{max}_{init}=0.3$ m.
  For each experiment, the sampling results from the multi-objective optimization are presented, and several structural examples are discussed in detail.
}%
{%
  本研究の実験設定について述べる.
  まず, 本研究では$N_{joint}=\{6, 7\}$と変化させて2つの実験を行う.
  評価関数計算に関するパラメータについては, 一般性と計算量を考慮して$x^{\min} = -0.5, x^{\max} = 0.5, y^{\min} = -0.5, y^{\max} = 0.5, z^{\min} = -0.1, z^{\max} = 0.5, d_{voxel} = 0.2, N_{rand} = 30$とした.
  その他のパラメータについては, 一般的なマニピュレータを考え, $m_{motor}=0.5$ kg, $\rho=1.0\times10^3$ kg/m$^3$, $r=0.015$ m, $L_{total}=0.6$ m, $L^{max}_{init}=0.3$ mとしている.
  各実験について, 多目的最適化におけるサンプリング結果を示し, いくつかの構造例を詳しく議論する.
}%

\begin{figure}[t]
  \centering
  \includegraphics[width=0.81\columnwidth]{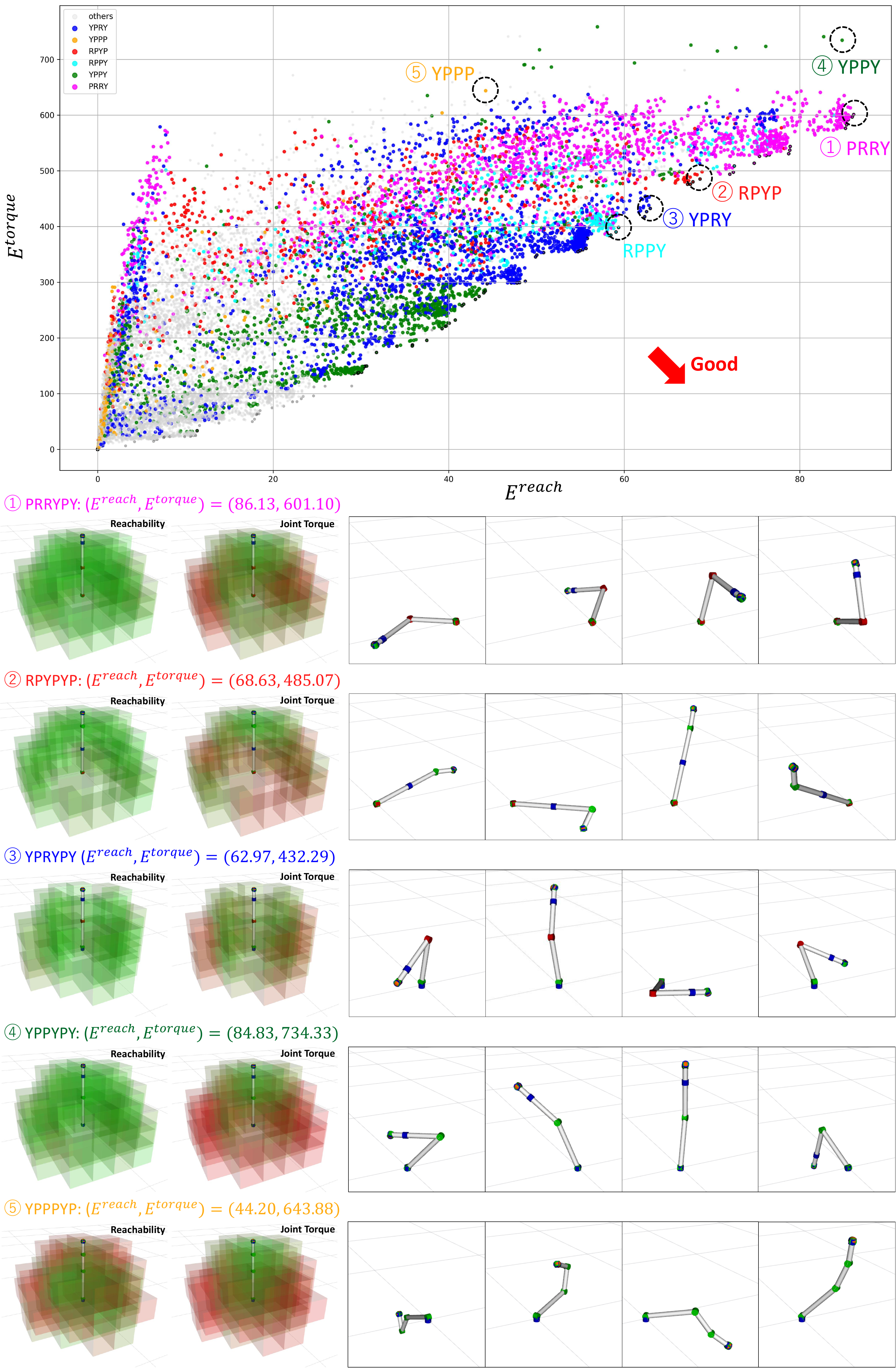}
  \caption{Optimization results for a 6-DOF manipulator: The upper figure visualizes the sampling results, color-coded based on differences in the first four joint types. The lower figure presents a subset of the solutions, showing end-effector reachability, joint torque, and four postures generated by assigning random joint angles.}
  \label{figure:exp-6dof}
\end{figure}

\subsection{Multi-Objective Optimization of 6-DOF Manipulator} \label{subsec:6dof-exp}
\switchlanguage%
{%
  The experimental results for $N_{joint}=6$ are shown in \figref{figure:exp-6dof}.
  In the sampling results, different colors represent variations in the first four joint configurations.
  Pareto-optimal solutions are highlighted with black circular frames.
  The six main configurations observed were \textcolor{magenta}{PRRY}, \textcolor{red}{RPYP}, \textcolor{blue}{YPRY}, \textcolor{cyan}{RPPY}, \textcolor{green}{YPPY}, and \textcolor{orange}{YPPP}.
  Among these, the joint configurations with the highest reachability within the Pareto-optimal solutions (\textcolor{magenta}{PRRY}, \textcolor{red}{RPYP}, and \textcolor{blue}{YPRY}) are illustrated in the lower figure of \figref{figure:exp-6dof}.
  Additionally, the joint configurations with the highest reachability among the sampled non-Pareto-optimal solutions (\textcolor{green}{YPPY} and \textcolor{orange}{YPPP}) are also presented.
  Since \textcolor{cyan}{RPPY} is nearly identical to \textcolor{magenta}{PRRY} due to symmetry along the $x$- and $y$-axes, it is omitted.

  A general trend observed is that, even within the same joint structure, increasing reachability often results in placing two DOFs near the origin (shoulder), one Pitch or Roll joint at the intermediate position (elbow), and three DOFs at the end-effector (wrist).
  This finding is consistent with what was reported in \cite{paden1988optimal}.

  The specific joint configurations are discussed below.
  \textcolor{magenta}{PRRY} exhibits both high overall reachability and low required torque.
  In particular, \textcolor{magenta}{PRRY} has low torque around the $y$-axis, while \textcolor{cyan}{RPPY} has low torque around the $x$-axis.
  \textcolor{red}{RPYP} places the Yaw joint in the middle, allowing the end-effector to achieve various postures.
  However, its reachability is reduced in positions close to the body.
  This structure reflects a conflict between two motives, resulting in an indecisive configuration.
  One aims to determine the 3D orientation with the initial RPY and place a Pitch joint in the middle.
  The other seeks to allocate all three RPY degrees of freedom to the end-effector.
  \textcolor{green}{YPPY} has two consecutive Pitch joints after a Yaw joint, making it susceptible to gravitational effects.
  However, its overall reachability remains quite high.
  This configuration is used in manipulators such as ALOHA, COBOTTA, and xArm 6.
  Manipulators mounted on quadruped robot platforms like Boston Dynamics' Spot \cite{bostondynamics2025spot} also share this structure.
  \textcolor{orange}{YPPP} is the configuration adopted by ARX and myCobot.
  Similar to \textcolor{green}{YPPY}, it is directly influenced by gravitational effects.
  With three consecutive Pitch joints, its reachability is relatively low, but it offers the simplest and most intuitive kinematics.
}%
{%
  $N_{joint}=6$のときの実験結果を\figref{figure:exp-6dof}に示す.
  サンプリング結果では, 最初の4つの関節配置の違いから色付けをしている.
  また, パレート解には黒い丸枠をつけている.
  主な配置は\textcolor{magenta}{PRRY}, \textcolor{red}{RPYP}, \textcolor{blue}{YPRY}, \textcolor{cyan}{RPPY}, \textcolor{green}{YPPY}, \textcolor{orange}{YPPP}の6つであった.
  パレート解となっている\textcolor{magenta}{PRRY}, \textcolor{red}{RPYP}, \textcolor{blue}{YPRY}の配置の中で最もreachabilityが高い関節配置と, パレート解となっていない\textcolor{green}{YPPY}や\textcolor{orange}{YPPP}のサンプリングの中で最もreachabilityが高い関節配置を実際に下図に示している.
  なお, \textcolor{cyan}{RPPY}は$x$軸方向と$y$軸方向に関する対称性から\textcolor{magenta}{PRRY}とほとんど同じであるため除いている.

  全体的な傾向として, 同じ関節構造の中でもreachabilityを増やそうとすると, 肩となる原点付近に2自由度, 肘となる中間付近にPitchまたはRollを1自由度, 手先に3自由度を配置することが多い.
  これは\cite{paden1988optimal}が示していることと同様である.

  以下具体的な関節配置について述べる.
  \textcolor{magenta}{PRRY}は全体的にreachabilityも高く, かつ必要なトルクも低い.
  特に, \textcolor{magenta}{PRRY}は$y$軸付近でのトルクが小さく, \textcolor{cyan}{RPPY}は$x$軸付近でのトルクが小さい.
  \textcolor{red}{RPYP}はYaw関節が中に来てしまっているので, 手先は様々な姿勢を取れるものの, 身体に近い位置でのreachabilityが下がっている.
  これは, 最初のRPYで3次元方向の狙いを定め, 中間にPitch関節を持っていきたいという思考と, 手先にRPYの3自由度を持っていきたいという思考がせめぎあい, どっちつかずとなっている.
  \textcolor{green}{YPPY}はYaw関節の後にPitch関節が2回並び, 重力の影響を受けやすい構造な一方で, 全体的なreachabilityはかなり高い.
  ALOHA, COBOTTA, xArm 6はこれと同じ構造をしている.
  また, Bosyton Dyanmics Spot \cite{bostondynamics2025spot}等に載るマニピュレータも同じ構造が多い.
  \textcolor{orange}{YPPP}はARXやmyCobotが採用している方式である.
  \textcolor{green}{YPPY}と同様に重力の影響をダイレクトに受けやすい.
  また, Pitch関節が3つ連続しているのもあり, reachabilityもあまり高くない.
  一方で, キネマティクスは最もシンプルで分かりやすい.
}%

\begin{figure}[t]
  \centering
  \includegraphics[width=0.81\columnwidth]{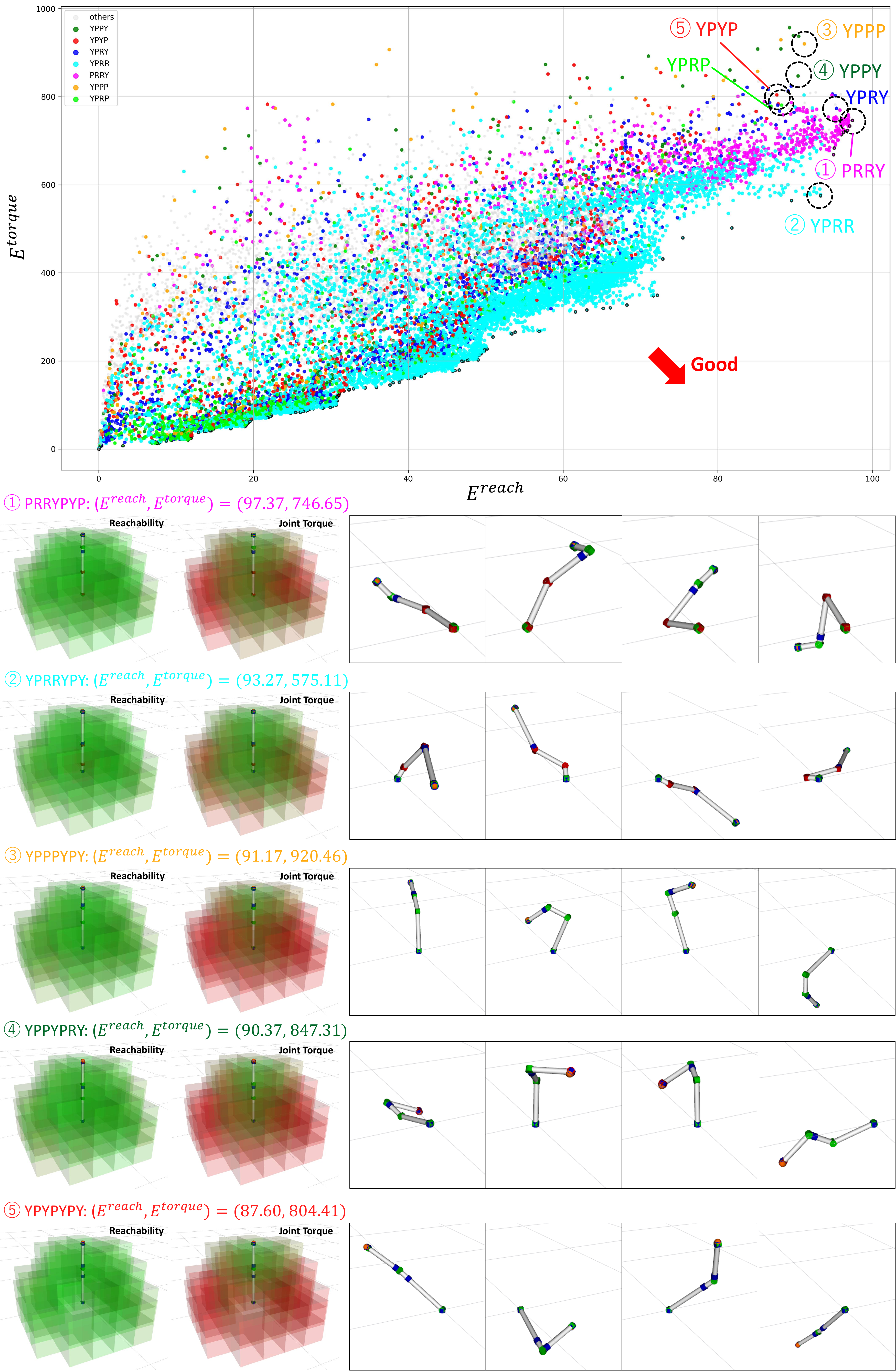}
  \caption{Optimization results for a 7-DOF manipulator: The upper figure visualizes the sampling results, color-coded based on differences in the first four joint types. The lower figure presents a subset of the solutions, showing end-effector reachability, joint torque, and four postures generated by assigning random joint angles.}
  \label{figure:exp-7dof}
\end{figure}

\subsection{Multi-Objective Optimization of 7-DOF Manipulator} \label{subsec:7dof-exp}
\switchlanguage%
{%
  The experimental results for $N_{joint}=7$ are shown in \figref{figure:exp-7dof}.
  The main joint configurations observed were \textcolor{magenta}{PRRY}, \textcolor{cyan}{YPRR}, \textcolor{blue}{YPRY}, \textcolor{orange}{YPPP}, \textcolor{green}{YPPY}, \textcolor{red}{YPYP}, and \textcolor{lime}{YPRP}.
  Note that the color assignments for the joint configurations may differ between the 6-DOF and 7-DOF cases.
  Among these, the joint configurations with the highest reachability within the Pareto-optimal solutions (\textcolor{magenta}{PRRY} and \textcolor{cyan}{YPRP}) are illustrated in the lower figure of \figref{figure:exp-7dof}.
  Additionally, the joint configurations with the highest reachability among the sampled non-Pareto-optimal solutions (\textcolor{orange}{YPPP}, \textcolor{green}{YPPY}, and \textcolor{red}{YPYP}) are also presented.

  Overall, the trends observed in the 7-DOF case are somewhat different from those in the 6-DOF case.
  While two degrees of freedom are concentrated near the origin (shoulder) and three degrees of freedom at the end-effector (wrist), as in the 6-DOF case, there are two distinct approaches for distributing the remaining two degrees of freedom: one where they are grouped around the elbow, dividing the entire linkage into a 1:1 ratio, and another where they are distributed more evenly in a 1:1:1 ratio.

  The specific joint configurations are discussed below.
  \textcolor{magenta}{PRRY} exhibits both high reachability and low required torque, consistent with the results from the 6-DOF case.
  As the degrees of freedom at the end-effector increase, reachability improves, but the required torque also increases.
  \textcolor{cyan}{YPRR} is a modification of the 6-DOF \textcolor{magenta}{PRRY} with an additional Yaw joint at the base.
  Since the first Yaw joint does not experience significant torque, the required torque remains nearly the same as in the 6-DOF \textcolor{magenta}{PRRY}, while reachability is improved.
  \textcolor{orange}{YPPP}, \textcolor{green}{YPPY}, and \textcolor{red}{YPYP} have simple and intuitive joint structures, but as in the 6-DOF case, are more directly affected by gravitational effects.
  All three exhibit similar performance, but among them, \textcolor{red}{YPYP} has the lowest torque requirements.
  \textcolor{red}{YPYP} is one of the most common structures in industrial robots, and manipulators such as Franka Emika Panda, Sawyer, and Kinova Gen3 adopt similar configurations.
  Additionally, a robot with the \textcolor{lime}{YPRP} structure is the Robai Cyton Gamma \cite{robai2025cytongamma}.
}%
{%
  $N_{joint}=7$のときの実験結果を\figref{figure:exp-7dof}に示す.
  主な配置は\textcolor{magenta}{PRRY}, \textcolor{cyan}{YPRR}, \textcolor{blue}{YPRY}, \textcolor{orange}{YPPP}, \textcolor{green}{YPPY}, \textcolor{red}{YPYP}, \textcolor{lime}{YPRP}の7つであった.
  パレート解となっている\textcolor{magenta}{PRRY}, \textcolor{cyan}{YPRP}の配置の中で最もreachabilityが高い関節配置と, パレート解となっていない\textcolor{orange}{YPPP}, \textcolor{green}{YPPY}, \textcolor{red}{YPYP}のサンプリングの中で最もreachabilityが高い関節配置を実際に下図に示している.
  全体的な傾向は6自由度と多少異なり, 原点付近に2自由度, 手先に3自由度が集中していることは同様だが, 残りの2自由度を一箇所に集めて肘を中心として全体リンクを1:1に分割する場合と, 一箇所に集めずに1:1:1に分割する場合がある.

  以下具体的な関節配置について述べる.
  \textcolor{magenta}{PRRY}は全体的にreachabilityも高く, かつ必要なトルクも低い, この特性は6自由度のときと一致している.
  一方6自由度のときに比べて, 手先に自由度がさらに増えるためreachabilityは増しているが, 同時に必要なトルクも上昇している.
  \textcolor{cyan}{YPRR}は6自由度のときの\textcolor{magenta}{PRRY}の手前にYawを追加した形である.
  最初のYaw関節にはトルクがかからないため, 必要なトルクは6自由度のときの\textcolor{magenta}{PRRY}とほぼ同じで, reachabilityだけ向上している.
  \textcolor{orange}{YPPP}, \textcolor{green}{YPPY}, \textcolor{red}{YPYP}はシンプルで分かりやすい関節構造を持つが, 6自由度のときと同様に重力の影響をダイレクトに受けやすい.
  どれもが似たような性能を示しているが, その中でも\textcolor{red}{YPYP}が最もトルクが小さい.
  \textcolor{red}{YPYP}は一般的な産業用ロボットに最も多く見られる構造で, Franka Emika PandaやSawyer, Kinova Gen3も同様の構造を持っている.
  他にも, \textcolor{lime}{YPRP}の構造を持つロボットとしてはRobai Cyton Gamma \cite{robai2025cytongamma}が存在している.
}%

\section{Discussion} \label{sec:discussion}
\switchlanguage%
{%
  The obtained results are summarized and discussed below.
  First, regarding the trends in the sampling results, 6-DOF manipulators showed significant differences in joint structure performance, with distinct characteristics in terms of reachability and torque.
  In contrast, 7-DOF manipulators exhibited relatively high reachability across different joint structures, with no significant differences observed.
  This highlights the importance of joint configuration in 6-DOF manipulators.
  Additionally, while the link length ratios were generally similar among 6-DOF manipulators, greater variation was observed in the 7-DOF case.

  Next, regarding specific joint configurations, many of the configurations found in the Pareto-optimal solutions were not commonly seen in real-world manipulators.
  In particular, the \textcolor{magenta}{PRRY} structure was found to have the highest reachability and the lowest torque for both 6-DOF and 7-DOF manipulators.
  This structure can be interpreted as a SCARA-type robot that moves horizontally, rotated using a Pitch joint.
  By concentrating the torque on the first Pitch joint and reducing the force applied to subsequent Roll joints, this configuration reduces the required torque by one Pitch joint compared to more conventional manipulators that connect to a YPP structure.
  In contrast, common manipulator structures such as \textcolor{orange}{YPPP} and \textcolor{green}{YPPY} are more susceptible to gravitational effects but feature simple kinematics, making their motion intuitive and easy to understand.
  While high gear ratios in traditional robots have mitigated these issues, the advancement of low-gear-ratio direct-drive motors may accelerate demand for structures that further reduce the effects of gravity in the future.

  In this study, multi-objective optimization was performed solely from the perspectives of end-effector reachability and joint torque.
  However, in reality, factors such as manufacturability and kinematic simplicity are also important.
  Considering these aspects to determine the most optimal structure would be highly meaningful.
  Rather than relying on human intuition, optimization based on data can provide interesting insights for future robot design.
  Additionally, another future challenge is determining the most optimal structure when incorporating not only rotational joints but also prismatic joints, closed-loop structures, and wire-driven mechanisms.
}%
{%
  これまで得られた結果をまとめ, 考察する.
  まずサンプリング結果の傾向であるが, 6自由度マニピュレータの場合は各関節構造の性能が大きく異なり, reachabilityやトルクの観点からも異なる特性が見られた.
  一方で, 7自由度マニピュレータの場合は, どの関節構造でもそれなりにreachabilityが高く, あまり大きな差が見られなかった.
  このことから, 6自由度マニピュレータの関節配置の重要性がよく分かる.
  また, 6自由度マニピュレータはリンクの長さ比率はどれも似ているが, 7自由度マニピュレータではここにバラエティが見られた.

  次に具体的な関節配置についてであるが, パレート解に存在する関節配置は, あまり実際のマニピュレータには見られないものが多かった.
  特に\textcolor{magenta}{PRRY}という構造が6自由度マニピュレータにおいても, 7自由度マニピュレータにおいても最もreachabilityが高く, かつトルクが小さいことが分かった.
  この構造は, 水平に動くスカラ型ロボットをPitch関節により回転させた構造と捉えることができる.
  かかるトルクを最初のPitch関節に集中させてその後のRoll関節にかかる力を減らすことで, YPPと繋げるような一般的なマニピュレータよりも, Pitch関節ひとつ分のトルクを削減することができる.
  これに対して, 一般的なマニピュレータに見られる\textcolor{orange}{YPPP}や\textcolor{green}{YPPY}といった構造は, 重力の影響は受けやすいものの, キネマティクスがシンプルで直感的に動きがわかりやすい.
  これまでのロボットはギア比が高くあまり問題にならなかったが, 現在のギア比が低いダイレクトドライブ系のモータの発展が進むと, 今後より重力の影響を軽減できる構造が求められる可能性はある.

  今後の展望について述べる.
  本研究では, 手先のreachabilityと関節トルクの観点のみから多目的最適化を行った.
  一方で, 本来は構造としての作りやすさや, キネマティクスとしてのシンプル差も重要であり, これらを考慮してどんな構造が最も最適なのかを考えることは非常に有意義である.
  人間の経験的な勘ではなく, データに基づいた最適化によって, 今後のロボットの設計に対する面白い示唆を与えることができると考えている.
  また, 回転関節のみならず, 直動関節や閉リンク構造, ワイヤ駆動等を組み合わせた時に, 最も良い構造とは何であるが, これも今後の課題である.
}%

\section{Conclusion} \label{sec:conclusion}
\switchlanguage%
{%
  In this study, a multi-objective optimization was performed from the perspectives of end-effector reachability and joint torque to analyze various manipulator structures.
  The design parameters consisted of discrete joint types and continuous link lengths, with additional constraints on joint sequences and total link length.
  A multi-objective optimization was efficiently conducted using the multivariate Tree-Structured Parzen Estimator (TPE).
  The optimization results revealed multiple novel manipulator structures that had not been previously explored while also clarifying the positions of existing manipulator structures within the optimization sampling.
  The differences in sampling results and Pareto-optimal solutions between 6-DOF and 7-DOF manipulators also provided interesting insights.
  Given the rapid advancements in motor technology, this study offers valuable guidance for future manipulator design.
  Of course, several interesting future challenges remain, such as the need to predefine certain parameters and the lack of consideration for kinematic simplicity or manufacturability.
  Nevertheless, this study provides a broad perspective on the performance of different joint structures.
}%
{%
  本研究では, 手先のreachabilityと関節トルクの観点から多目的最適化を行い, 多様なマニピュレータの構造について議論した.
  設計パラメータを関節の種類とリンク長という離散的かつ連続的なパラメータとし, それらについて関節順序とリンク全長の制約を加えたうえで, 効率の良いmultivariate Tree-Structured Parzen Estimator (TPE)を用いた多目的最適化を行った.
  最適化結果から, これまでに見ないマニピュレータの構造が複数得られると同時に, 既存のマニピュレータの構造が最適化サンプリングの中でどのような位置にあるのかを明らかにした.
  6自由度と7自由度でのサンプリング結果の違いやパレート解の違いからも, 興味深い結果が得られている.
  本研究は, モータ技術の発展著しい現在に置いて, 今後のマニピュレータ設計に重要な示唆を与えるものであると考えている.
  もちろん事前にいくつかのパラメータを設定していたり, キネマティクスのシンプルさや作りやすさを考慮していなかったりと, いくつか面白い今後の課題が残されているが, それでも, 関節構造の性能に関する全体像は示すことが出来たのではないだろうかと考えている.
}%

{
  \bibliographystyle{junsrt}
  \bibliography{main}
}

\end{document}